\documentclass[nonacm, manuscript]{acmart}
\AtBeginDocument{%
  }

\usepackage{amsmath}
\usepackage{enumitem}
\usepackage{listings}
\usepackage{subcaption}
\usepackage{caption}
\setlist[description]{leftmargin=1.8em, labelsep=0.6em, itemsep=2pt, topsep=2pt}

\begin{document}

\title{Translating Cultural Choreography from Humanoid Forms to Robotic Arm}

\author{Chelsea-Xi Chen}
\authornote{Both authors contributed equally to this research.}
% \email{Xi.Chen2404@student.xjtlu.edu.cn}
\orcid{0009-0006-7607-1499}
\affiliation{%
  \institution{Xi'an Jiaotong-Liverpool University}
  \city{Suzhou}
  \country{P.R. China}}

\author{Zhe Zhang}
% \email{Yian.Tang24@student.xjtlu.edu.cn}
% \orcid{1234-5678-9015}
\authornotemark[1]
\affiliation{%
  \institution{Xi'an Jiaotong-Liverpool University}
  \city{Suzhou}
  \country{P.R. China}}

\author{Aven-Le Zhou}
\orcid{0000-0002-8726-6797}  
\email{aven.le.zhou@gmail.com}
\authornote{Corresponding Author.}
\affiliation{%
  \institution{Xi'an Jiaotong-Liverpool University}
  \city{Suzhou}
  \country{P.R. China}}

\renewcommand{\shortauthors}{Chen et al.}

\begin{abstract}
Robotic arm choreography often reproduces trajectories while missing cultural semantics. This study examines whether symbolic posture transfer with joint space compatible notation can preserve semantic fidelity on a six degree of freedom arm and remain portable across morphologies. We implement ROPERA, a three stage pipeline for encoding culturally codified postures, composing symbolic sequences, and decoding to servo commands. A scene from Kunqu opera, \textit{The Peony Pavilion}, serves as the material for evaluation. The procedure includes corpus based posture selection, symbolic scoring, direct joint angle execution, and a visual layer with light painting and costume informed colors. Results indicate reproducible execution with intended timing and cultural legibility reported by experts and audiences. The study points to non anthropocentric cultural preservation and portable authoring workflows. Future work will design dance informed transition profiles, extend the notation to locomotion with haptic, musical, and spatial cues, and test portability across platforms.
\end{abstract}



\maketitle

\section{Introduction}
Robotic performance art has gained momentum with advances in computational choreography, kinematic control, and embodied robotics \cite{ref1,ref2}. Over the past decade, two main technical trajectories, humanoid robot choreography and industrial robotic arm-based choreography (RAC)\cite{ref3,ref15,ref35,ref43}, have emerged\cite{ref4,ref44,ref45,ref46}. These systems typically focus on mapping human spatial trajectories, synchronizing rhythmic structures, and coordinating multi-axis movements to achieve expressive robotic performances. Among them, RAC has become the predominant form in live performances and installation contexts due to its precision, repeatability, and mechanical stability, which are rooted in industrial automation frameworks \cite{ref2, ref3, ref4}.

Despite these advances, trajectory replication alone does not capture the symbolic and semantic layers that characterize culturally grounded choreography. Codified dance traditions such as Kunqu opera and Noh theatre encode gestural grammar, rhythmic nuance, and semantic structure across generations \cite{ref1,ref5}. In such settings, bodily movement functions as a structured semiotic system rather than a purely motor output. As robotic systems move into performative spaces, an open question is how they can engage with this symbolic depth without relying on anthropomorphic imitation.

This study treats choreography as a symbolic translation across bodies, morphologies, and materials, rather than solely as spatial motion planning \cite{ref3,ref6}. The framing emphasizes gesture as interface and foregrounds meaning-making over mimicry. Prior work in non-anthropomorphic choreography suggests that expressive intent can emerge through modular grammar, spatial encoding, and gesture-based interaction systems \cite{ref4,ref7}. Building on these insights, we explore culturally situated robotic performance with an emphasis on symbolic representation and servo-level realization.

\subsection{Background}
Recent advances in machine learning, trajectory optimization, and neural network-based motor control have improved the smoothness and continuity of robot arm movement and trajectories \cite{ref8,ref9,ref10,ref11}. Even with this progress, RAC still struggles to achieve kinematic control with rhythmic nuance and embodied dynamics comparable to those of expert human dancers. Systems often fall short of reproducing the complex kinetic layering, organic energy transfer, and full-body coordination that underpin human expressivity\cite{ref9,ref12}. In practice, RAC remains early relative to human codified choreography. Human codified choreography refers to movement frameworks within cultural or classical traditions that rely on standardized, named, and historically transmitted posture vocabularies and compositional structures. Several constraints continue to limit the aesthetic and cultural potential of robotic arm performance.

A recurring limitation concerns cultural expressivity. Dance, as a highly codified form of embodied knowledge, encodes symbolic postures, ritualized patterns, and culturally specific pose vocabularies \cite{ref6, ref13, ref14}. Unlike human choreographic traditions that accumulate reusable knowledge and symbolic forms over centuries, RAC work is frequently built from scratch for each performance. This leads to limited scalability, fragmented design procedures, and shallow cultural depth \cite{ref12,ref15}. Current RAC also relies heavily on low-level kinematic imitation. Sensitivity to cultural context is often lacking, which can result in decontextualized or mechanically repetitive motion \cite{ref16}. The lack of anatomical correspondence further complicates faithful translation of symbolic postures to robotic embodiment. At the same time, research in architectural robotics suggests that non-anthropomorphic morphologies can still achieve gestural intelligibility through posture abstraction\cite{ref3}.

Another limitation is the absence of an efficient symbolic notation frame for RAC. Development is hindered by the lack of structured choreography representation and by the absence of a symbolic posture vocabulary within a unified symbolic notation framework. Established notations for human choreography, including Labanotation and Benesh Movement Notation, are tightly coupled to human anatomy, spatial orientation, and weight shift mechanics \cite{ref17,ref18}. These systems do not map directly to robotic joint space control, especially for multi-degree-of-freedom arms \cite{ref3,ref12,ref13,ref14}. As a result, RAC lacks an integrated path that links cultural posture representation to joint space execution.

\subsection{ROPERA framework}
To address these challenges, this work adopts a symbolic posture transfer paradigm \cite{ref1,ref4,ref6}, in which culturally codified postures are transformed into symbolic motion primitives that are independent of the embodiment morphology. Unlike humanoid robotic choreography aiming at anatomical replication, this symbolic posture transfer emphasizes the encoding of choreographic semantics rather than biomechanical imitation. These symbolic posture units are directly mapped onto servo-level control parameters, enabling culturally grounded performances across diverse robotic morphologies.

This approach targets performance genres characterized by highly codified postural vocabularies, such as Kunqu Opera, Ballet, and Noh Theater, where semantically rich bodily movements in modular compositional structures offer a robust foundation for symbolic notation frameworks \cite{ref1,ref4,ref6}. Accordingly, we propose \textit{ROPERA}, a framework that translates codified cultural choreography into robotic arm–based choreography (RAC). The framework integrates three stages: the construction of a symbolic posture vocabulary mapped into joint-space-compatible robotic representations, the symbolic encoding of repertoire choreography into symbolic sequences with temporal alignment, and servo-level execution that decodes symbolic notation into direct joint-space control commands for robotic choreography.

\subsection{Kunqu Opera and Peony Pavilion}
As a representative case study for systematic validation, this study selects Kunqu opera, a classical Chinese performing art form originating in the 16th century, whose choreography is systematically codified through the “hand-eye-body-method-step” framework \cite{ref5, ref7, ref19, ref20, ref21, ref22}. Kunqu’s modular composition, symbolic expressivity, and structural repetition make it particularly suitable for defining motion primitives, enabling the symbolic abstraction of posture and modeling robotic embodiment.

We adapt \textit{The Peony Pavilion}, a canonical work of Kunqu opera, to demonstrate the framework and to showcase both the technical feasibility and aesthetic expressiveness of culturally grounded robotic performance. Kunqu opera and \textit{The Peony Pavilion} serve as a validation case in this study, and the \textit{ROPERA} framework is designed to generalize to other highly structured and culturally codified dance and choreographic genres.

This study presents the ROPERA framework, which translates culturally codified humanoid choreography into servo-executable robot arm choreography (RAC), a symbolic notation compatible with joint space for robotic arms, with a path to non-humanoid platforms. It also features an implemented RAC performance, accompanied by an evaluation that demonstrates cultural expressivity, technical feasibility, and aesthetic potential. 

\section{Related Work}
Robotic arms have become central to performance-oriented choreography due to their high-precision control, stability, and operational flexibility \cite{ref3,ref4}. While recent advances in machine learning and trajectory optimization enhance spatial smoothness and rhythmic synchronization \cite{ref9,ref23}, these methods often prioritize physical parameters and overlook cultural or semantic integration \cite{ref22,ref24}. To address these constraints, recent work shifts from trajectory imitation to culturally grounded symbolic notation frameworks \cite{ref1,ref6}. We examine four domains that shape this shift, including motion aesthetics and expressivity, cultural posture translation and symbolic abstraction, notation for robotic posture encoding, and systemic limitations in current RAC pipelines.

\subsection{Motion Aesthetics and Expressivity}
Expressive choreography depends on more than mechanical precision. It involves kinetic layering, rhythmic nuance, and the transfer of energy across the body. These qualities remain challenging to achieve in robotic arms, which often appear mechanical or segmented in their motion \cite{ref4,ref23}. While trajectory optimization methods can align motion with rhythm, they frequently fail to capture higher-level posture dynamics and choreographic intent \cite{ref9}. Emerging approaches aim to bridge this gap by utilizing frameworks such as Laban Movement Analysis to translate qualitative motion parameters into robot instructions \cite{ref25}. Aesthetic mapping frameworks also explore how choreographic principles can inform robotic control \cite{ref1, ref26}, However, the integration of these methods into robotic arm–based choreography (RAC) pipelines remains limited, often due to the absence of formalized symbolic representation.

Recent studies extend these efforts by exploring gesture not merely as motion output, but as an expressive and interpretable interface. For instance, Van der Linden et al. demonstrate how architectonic robotic forms can communicate semantic intent through structural posture and movement, treating gesture as a legible symbolic interface \cite{ref4}. Similarly, expressivity-driven models also regard gesture as a primary semantic unit and prioritize affective resonance over motor replication \cite{ref24}. These directions are reinforced by research that frames affective descriptors such as flow, weight, and effort as symbolic scaffolds in embodied interaction design \cite{ref27}. Wiberg et al. further argue that material and temporal dynamics contribute intrinsically to symbolic expressivity in movement-based interfaces \cite{ref28}. Building on this line, work on mechatronic expression proposes that symbolic form can be performatively meaningful beyond emotional realism, which challenges prevailing paradigms of expressivity in robotics \cite{ref29}.

\subsection{Cultural Posture Translation and Symbolic Abstraction}
Cultural choreography in genres such as Kunqu opera and Noh theater encodes symbolic postures in ritualized sequences \cite{ref6,ref30}, yet many RAC pipelines emphasize low-level kinematics and lack semantic awareness \cite{ref15}. Studies in human-robot interaction suggest that incorporating culturally specific postures can enhance both audience comprehension and engagement \cite{ref31}. Design-oriented HCI reaches a similar conclusion by showing that culturally embedded symbolic structures, such as weaving grammars, can operate as visual-semantic systems for interaction encoding \cite{ref32}. Research in embodied interaction demonstrates that cultural motifs and symbolic structures, including textile grammars and spatial gestural forms, can serve as programmable modules for expressive interaction \cite{ref4,ref32}. Real-time gesture frameworks developed for improvisational dance performances further highlight the potential of affective responsiveness in robotic choreography \cite{ref33}.

Even with these advances, few RAC frameworks offer a structured method for converting codified cultural postures into transferable robotic representations. Our work introduces symbolic posture abstraction that seeks to preserve semantic integrity while enabling servo-level control, with the goal of facilitating culturally meaningful robotic execution.

\subsection{Notation frameworks for Robotic Posture Encoding}
Traditional symbolic notation frameworks, such as Labanotation and Benesh movement notation, provide comprehensive frameworks for documenting human choreography; however, they are incompatible with servo-based robotic control due to their anatomical dependence \cite{ref17, ref18}. Adapting such frameworks for robotic platforms requires intermediate abstraction layers that reconcile cultural semantics with hardware-specific control structures.

Recent studies have proposed multi-modal notation frameworks for robotic dance\cite{ref34,ref35}, yet these primarily emphasize trajectory-level encoding rather than symbolic postural semantics. Computational surveys also highlight the absence of unified frameworks linking cultural modeling to joint-level execution \cite{ref36,ref37}.  Other efforts have explored symbolic robotic gesture systems detached from human anatomy, utilizing mobile and wearable robots to express gesture logic and modular composition through movement primitives \cite{ref4, ref7}.

To address these gaps, the ROPERA framework proposes a joint space-compatible symbolic notation that encodes culturally codified postures in servo-executable formats. The design aims to formalize symbolic modeling for culturally meaningful postures, to integrate symbolic encoding with joint-level execution, and to improve extensibility beyond specific platforms or performance contexts.

\begin{figure}
    \centering
    \includegraphics[width=1\linewidth]{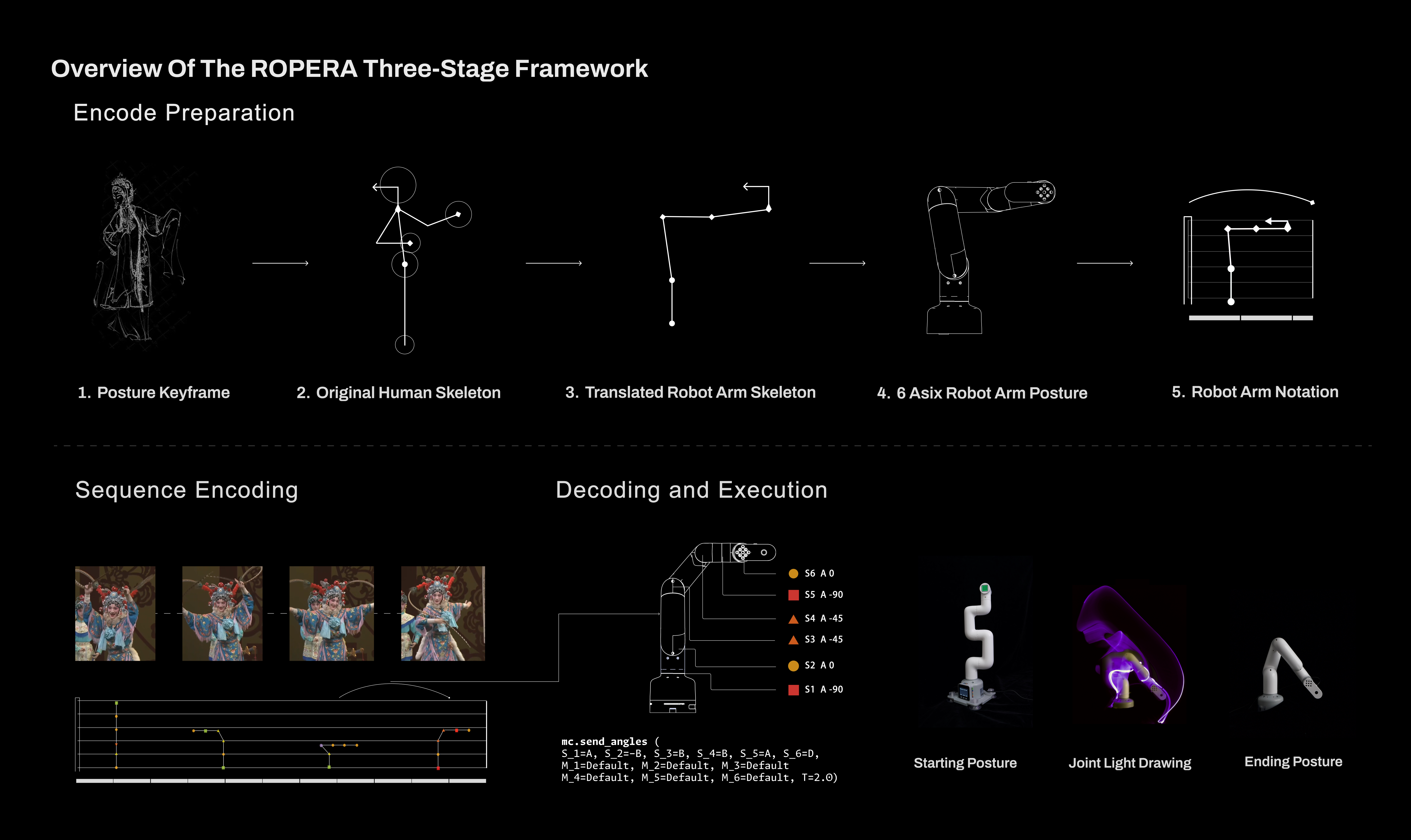}
    \caption{Overview of the ROPERA three-stage framework}
    \label{fig:placeholder}
\end{figure}

\begin{figure}
    \centering
    \includegraphics[width=1\linewidth]{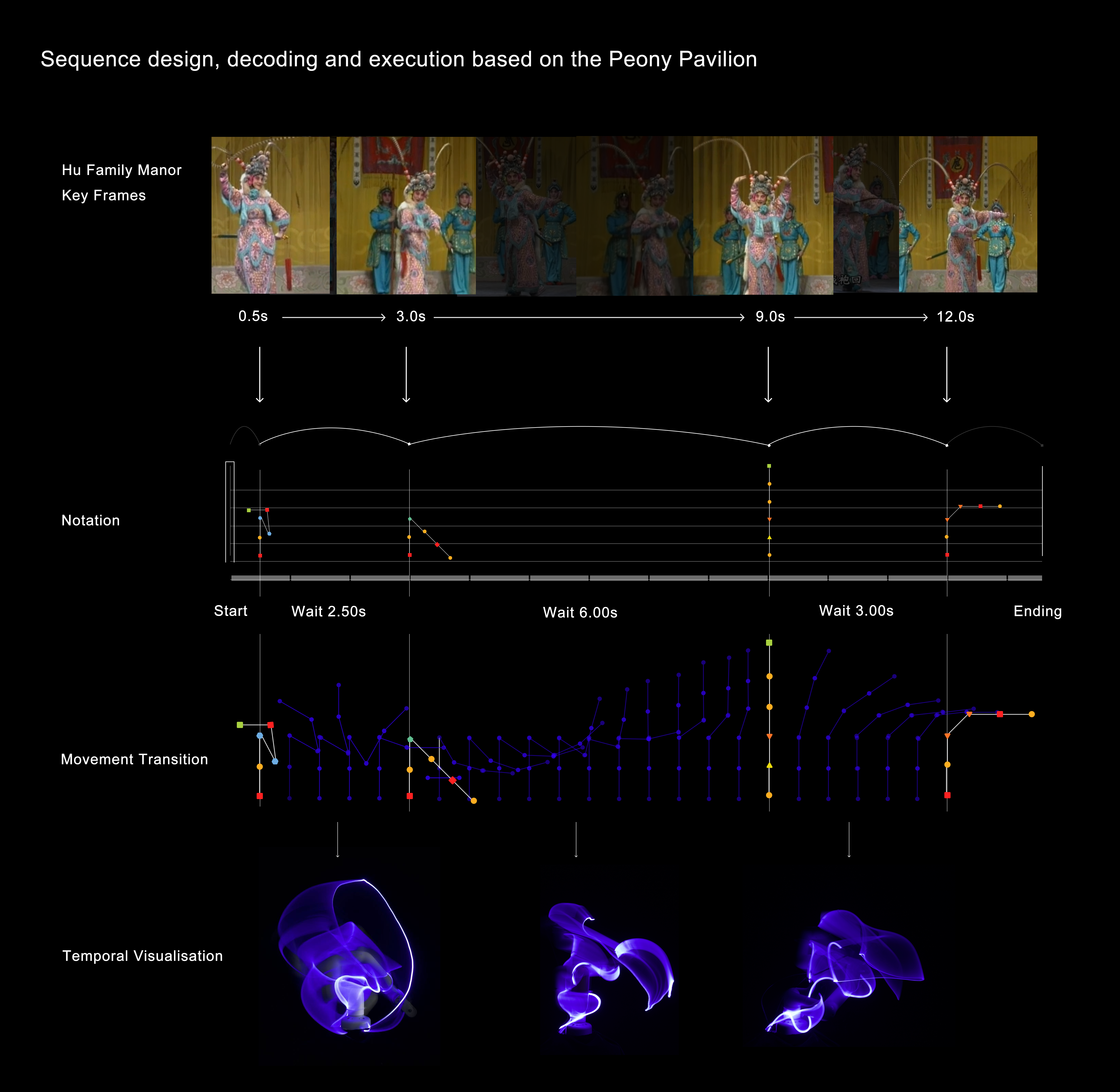}
    \caption{Sequence design, decoding and execution based on \textit{The Peony Pavilion}}
    \label{fig:placeholder}
\end{figure}

\section{ROPERA Framework Overview }
ROPERA targets cross-platform adaptability, and the current implementation is built and evaluated on a six-degree-of-freedom robotic arm as an example. ROPERA operates through a structured three-stage pipeline that seeks to bridge the semantic gap between cultural choreography and joint space robotic execution. As depicted in Figure 1, Encoding Preparation, Sequence Encoding, and Decoding and Execution operate together as an integrated pipeline that converts cultural choreography into stable, repeatable, and semantically expressive robotic performances. The framework addresses cultural and technical challenges identified in prior research on cultural pose transfer, symbolic motion decoding and encoding, and robotic performance \cite{ref1, ref4, ref6}.

\subsection{Encoding Preparation}
Encoding Preparation compiles a genre-specific vocabulary of symbolic human poses for robotic translation. The source corpus draws on two streams, recorded interviews with theatre experts \cite{ref41, ref42} and canonical treatises that document posture atlases and choreographic rules \cite{ref38, ref39, ref40}. From these materials, we identify culturally codified pose units. The resulting poses are treated as semantic primitives rather than motion templates, functioning as modular units of embodied meaning within the choreographic framework.

Each human pose is then transformed through a functional topological mapping into a mechanically feasible robotic configuration. Instead of replicating anatomical joint structures, the mapping abstracts choreographic meaning, including directional energy, stance hierarchy, and symbolic salutation, into robot-compatible motion constructs. The resulting robotic configurations, or robotic poses, form the symbolic foundation of the robotic posture vocabulary.

To enable platform-level encoding and sequential design, these robotic poses are recorded within a symbolic notation framework designed for multi-degree-of-freedom robotic arms. This notation discretizes joint-space states into symbolic units that are human-readable, servo-compatible, and temporally definable. The abstraction enables choreography to be transferred across robotic arms with different morphologies by reassigning servo mappings without altering symbolic logic.

\subsection{Sequence Encoding}
Sequence Encoding translates repertoire-specific choreography into a temporally ordered sequence of symbolic robotic poses. Drawing on the genre-specific posture vocabulary, key postural moments are extracted from canonical choreography through score annotation, motion segmentation, and expert-informed analysis. Each posture receives temporal metadata that specifies duration and structural role.

Using the topological mapping from the previous stage, annotated human postures are systematically converted into robotic configurations encoded in the symbolic notation framework. This encoding retains cultural semantics and embeds timing information, allowing for smooth modulation and seamless transitional dynamics. The symbolic sequence serves as a servo-compatible blueprint for choreography, structured to support rhythmic, semantic, and compositional coherence. Its symbolic nature allows for high-level operations, such as phrase remixing and genre-specific adaptation, to be performed before execution. As an ongoing effort, we are refining the granularity of temporal markers and transition flags, and we plan to compare alternative segmentation policies with expert feedback.

\subsection{Decoding and Execution}
Decoding and Execution convert symbolic choreography into executable joint space commands for the robotic arm. Each symbolic frame includes posture symbols together with temporal annotations and is decoded into servo-specific joint angles and transition parameters. The decoding maps symbolic values to concrete actuator positions, incorporating pose duration and inter-frame intervals to calculate velocity and acceleration profiles. Instead of relying on custom planning algorithms, the framework uses the robot’s built-in trajectory controller for smooth, time-synchronized motion transitions.

This decoding process yields a time-indexed stream of control instructions that preserves the spatial composition and timing of the original choreography. By separating symbolic design from platform-specific execution, portability is enabled through remapped servo assignments while keeping the symbolic structure unchanged. We are measuring timing accuracy, path smoothness, and mapping fidelity on a six-degree-of-freedom arm, and we plan to extend tests to additional morphologies to evaluate cross-platform transfer.

\section{Experiment: From Peony Pavilion to Robotic Arm Choreography}
This section demonstrates a full-cycle translation of the ROPERA framework using a segment from \textit{The Peony Pavilion}, a canonical Kunqu opera work. Within the three-stage architecture, we show encoding preparation, sequence design, and decoding and execution. Fig. 2 details the symbolic-to-servo pipeline and choreography decoding workflow. Beyond symbolic motion encoding and servo-level choreography realization, the implementation emphasizes the integration of visual elements inspired by Kunqu's costume aesthetics. Specifically, guided by traditional Kunqu attire, dominant palette motifs such as celadon green, peony pink, and ivory white are applied to the robot’s visible components and to light painting outputs.

\subsection{Kunqu Opera Encoding Preparation: Genre-based Pose Vocabulary and Mapping}

\subsubsection{Data Collection and Motion Primitive Extraction}

We constructed the corpus from several demonstration videos [38,39], a movement notation catalogue [40], and interview recordings with master performers [41,42]. Using these sources, we applied a unified annotation rule that treats a posture unit as a temporally stable configuration separable from adjacent transitions by an observable inflection in joint orientation.

From this corpus, we identified fifteen motion primitives, which are basic Kunqu posture units drawn from body stances and arm–sleeve techniques, normalized to a symbolic vocabulary for robotic mapping, that are illustrated in Fig. 3, including twelve upper-limb symbolic poses (e.g., sleeve lift, arm cross, shoulder pivot) and three full-body configurations (e.g., salutation stance, crouch, neutral posture). These primitives were selected based on frequency, structural significance, and cultural symbolism, and serve as the atomic units in later symbolic choreography composition. 

\begin{figure}
    \centering
    \includegraphics[width=1\linewidth]{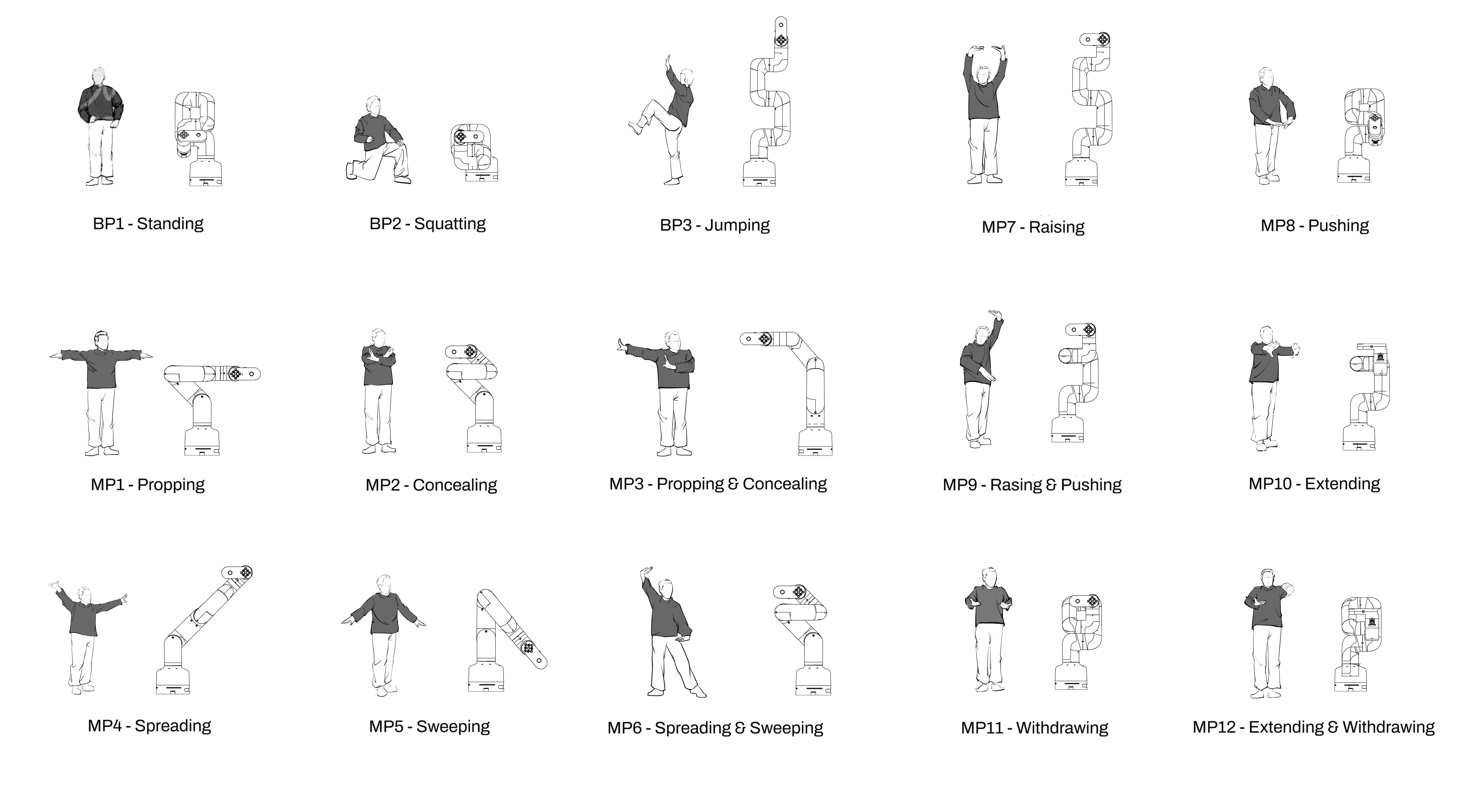}
    \caption{15 motion primitives in Kunqu Opera}
    \label{fig:placeholder}
\end{figure}

\subsubsection{Functional Topological Mapping to Robotic Arm Platform}
Given the morphological disparity between human anatomy and a six-degree-of-freedom robotic arm, direct anatomical mapping is infeasible. We therefore apply a functional topological mapping strategy in which each human body segment is associated with a robotic joint channel according to its role in producing spatial transformations.

In this model, the 6 servos are labeled s1–s6 from base (proximal) to end-effector (distal).  Fig. 4 summarizes functional roles for mapping choreographic meaning onto robotic articulation. Servos 1 and 2 primarily control base rotation and vertical elevation, and are mapped to head directionality and overall body orientation. Empirical tests showed that angular shifts in these axes effectively simulate human head tilts and full-body reorientation, contributing to the expressive framing of postures. Servo 3 serves as a shoulder-height pivot, mediating the spatial “stance” height. While not directly analogous to human joints, it functions as a vertical adjuster, supporting changes in limb elevation and center of gravity. Servos 4, 5, and 6 constitute the main articulation chain for simulating arm postures, which are central to Kunqu's symbolic vocabulary. These joints control forearm extension, wrist positioning, and palm orientation, respectively.

\begin{figure}
    \centering
    \includegraphics[width=1\linewidth]{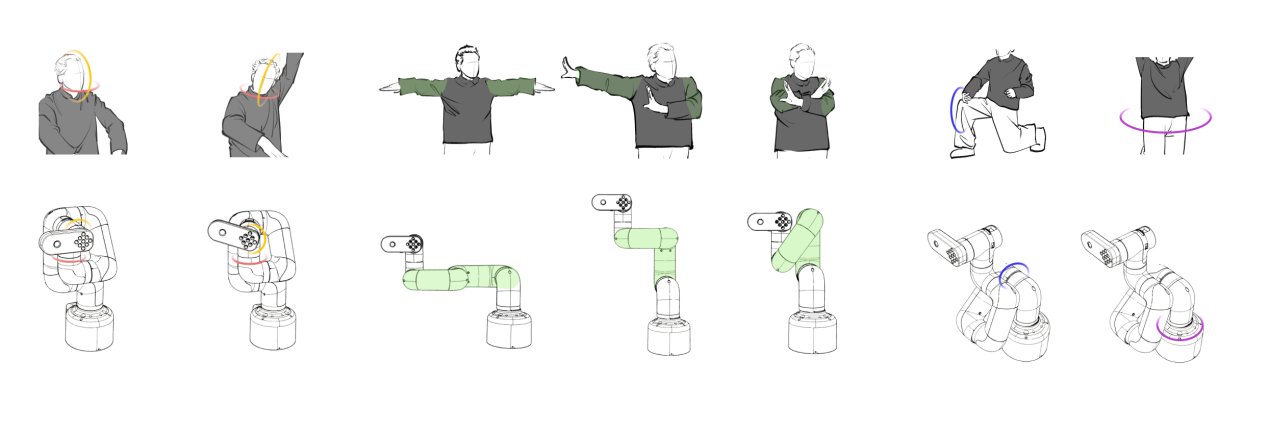}
    \caption{Joint mapping strategy from human to robot for head, gesture, height of center of mass, and rotation}
    \label{fig:placeholder}
\end{figure}

Due to base constraints and mechanical design, Servos 5 and 6 are mechanically coupled, meaning their rotation affects both vertical height and horizontal orientation of the entire upper limb. A compensatory control algorithm offsets interaction between Servo 5 and 6 to ensure posture stability when postures involve both rotational and vertical variation, such as crouching with extended arms. This function-based mapping preserves the semantic spatial logic of Kunqu postures while maintaining the expressive fluidity of robotic performance. Unlike anatomical replication, the approach employs functional topological mapping, associating each human body segment with a robotic joint channel in accordance with its role in generating spatial transformations.

\subsection{\textit{The Peony Pavilion} Sequence Encoding}
Building upon the symbolic posture vocabulary established in the encoding phase, a structured notation schema was developed to formally represent pose sequences in a format suitable for robotic execution. This schema facilitates high-level choreographic composition while maintaining direct compatibility with servo-level control. Each notation frame includes a discrete configuration of symbolic postures tied to specific servo angle states, a temporal annotation that controls pose duration and inter-frame rhythm, and a motion state flag that marks whether each joint remains static or transitions dynamically within the frame.

To demonstrate the schema’s operational viability, a twelve-frame symbolic choreography was composed based on a selected segment from \textit{The Peony Pavilion}, prioritizing emblematic transitions with strong cultural symbolism such as sleeve sweeping, salutation, and arm retraction. Postural choices follow expert knowledge and structural analyses of Kunqu conventions \cite{ref40}, so that the sequence captures emblematic patterns such as sleeve sweeping, salutation, and arm retraction.

The symbolic sequence was encoded using a symbolic notation framework structured as:
\begin{equation}
\text{Notation}=\bigl\{(S_1,\ldots,S_n),\ (M_1,\ldots,M_m),\ T\bigr\}.
\label{eq:notation}
\end{equation}, and it is also shown in Fig. 5 (a) where

\begin{description}
  \item[\(S\)] \(\{A,\ldots,I\}\) symbolic joint–angle bins at \(45^\circ\) increments.
  \item[\(M\)] motion state that indicates whether each joint changes in the current frame.
  \item[\(T\)] frame duration in seconds that sets rhythmic structure.
\end{description}

Each servo range is quantized into eight bins, A to I, in 45-degree steps. Bins are visually coded with distinct shapes and colors. The polygon edge count grows with the angle, so each additional edge indicates one more 45-degree increment. Fig. 5 (b) shows the codebook and a sample frame. Although bin E denotes 180 degrees, the platform reaches only ±175 degrees in practice, so values are clipped to this limit during execution.

\begin{lstlisting}[language=Python, caption={Example execution call}, label={lst:exec}]
mc.send_angles(S_1="A", S_2="B", S_3="H", S_4="H", S_5="A", S_6="D",M_1="Default", M_2="Default", M_3="Default",
    M_4="Default", M_5="Default", M_6="Default", T=2.0)
\end{lstlisting}

\begin{figure*}[t]
  \centering
  \begin{subfigure}{0.45\linewidth}
    \centering
    \includegraphics[width=\linewidth]{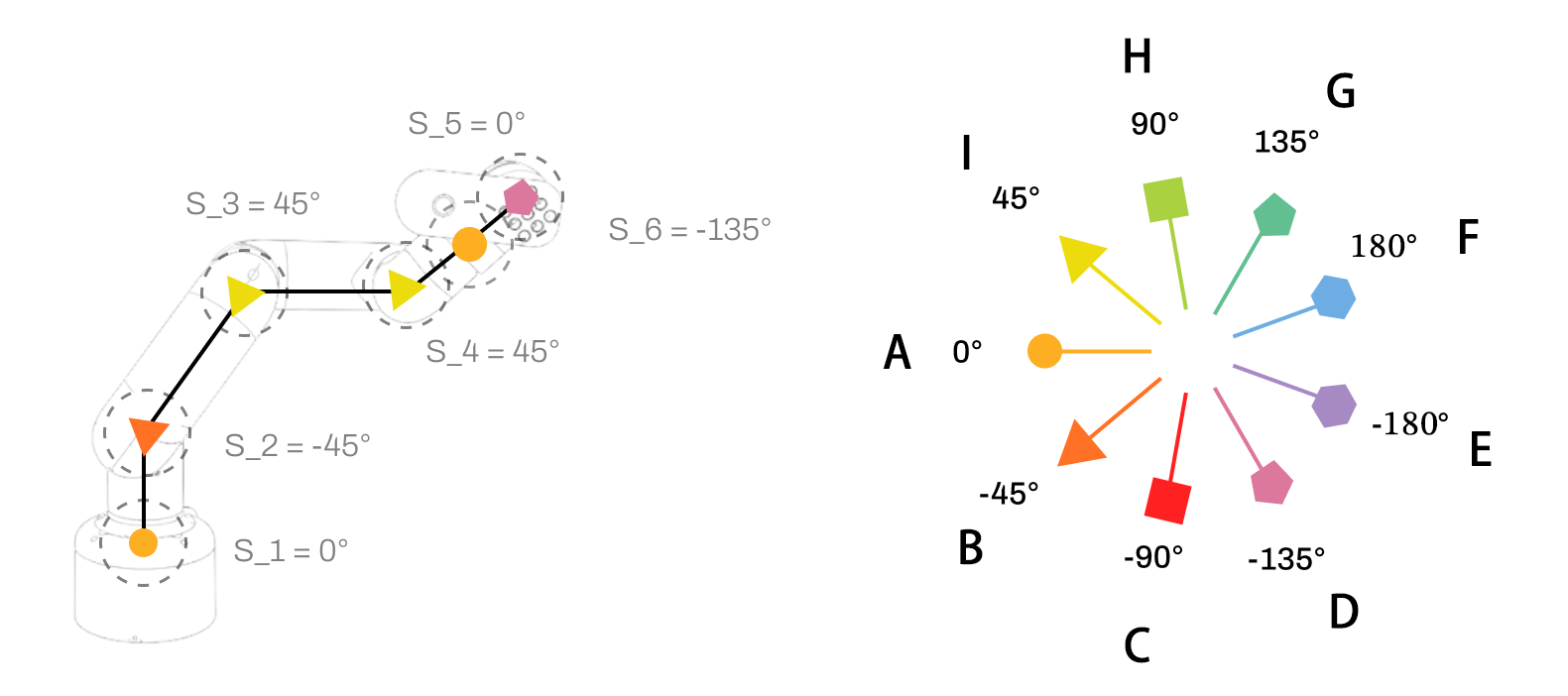}
    \caption{Notation overview}
  \end{subfigure}\hfill
  \begin{subfigure}{0.45\linewidth}
    \centering
    \includegraphics[width=\linewidth]{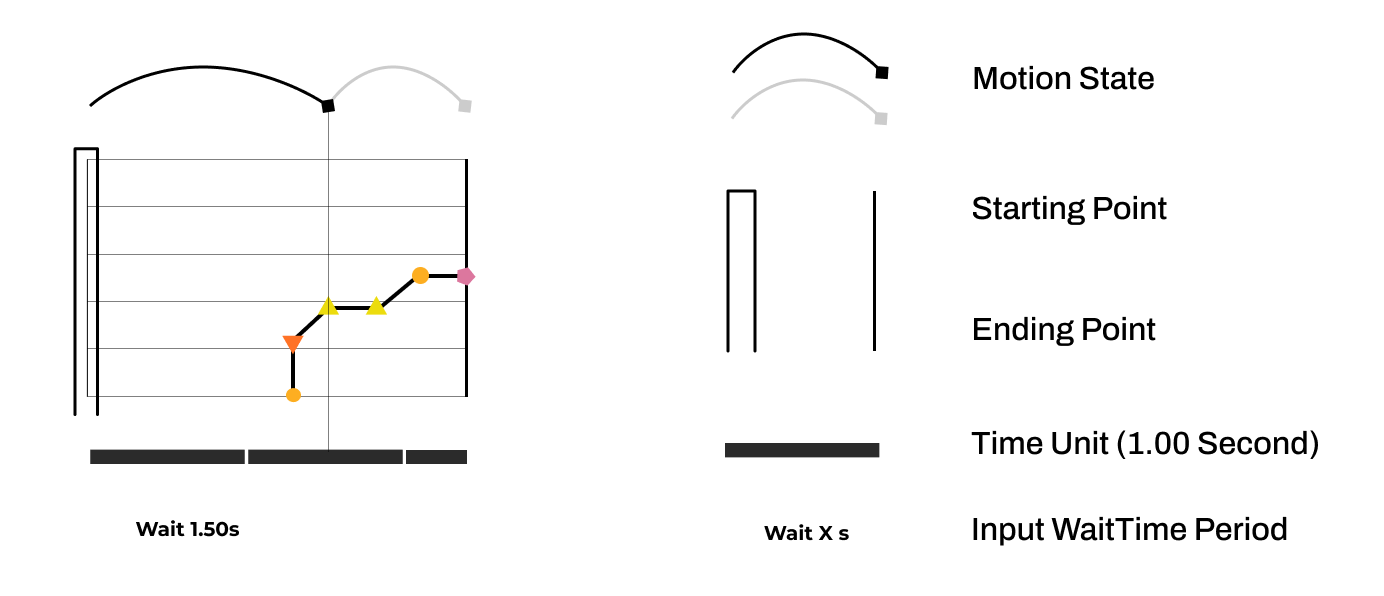}
    \caption{Symbolic frame execution and angle bins}
  \end{subfigure}
  \caption{Symbolic notation and execution corresponding to Eq.~\ref{eq:notation} and Listing~\ref{lst:exec}.}
  \label{fig:notation}
\end{figure*}

This symbolic encoding supports high-level composition, efficient editing, and transfer across platforms that differ in kinematic configuration. In the current implementation, with six servos (i.e., N = 6), the notation directly corresponds to the arm configuration. However, this framework remains architecture agnostic. By redefining the dimensionality N and the servo mapping rules, it can be generalized to robotic platforms with different morphologies and degrees of freedom, including snake-like manipulators, continuum arms, quadrupeds, and multi-legged machines. As a Work in Progress, we plan to compare remapping strategies across platforms and to gather expert feedback on notation granularity and mapping fidelity.

\subsection{Robot Arm Decoding and Execution}

The symbolic choreography encoded in earlier stages was actualized by a dedicated execution engine built on the native Python API of the \textit{MyCobot} robotic platform. Each symbolic frame includes discretized posture symbols, motion state flags, and temporal annotations, and is decoded into executable joint commands. Symbolic servo values, for example, A through H, are translated to angle values through predefined lookup tables. The framework selectively updates only those servos marked as dynamic in each frame, while static servos retain their prior configuration, thus enhancing frame-to-frame stability.

The engine operates under a joint angle control paradigm, allowing for the direct specification of servo angles and thereby bypassing the need for complex inverse kinematics or end-effector path planning. Frame transitions are scheduled using the temporal parameter T embedded within the notation schema, ensuring rhythmic synchronization and maintaining choreographic timing fidelity. To further stabilize motion output, servo transitions were implemented using fixed-angle step sizes combined with smoothing interpolations, which minimized jitter and improved repeatability. In our experiment, we adapted the default trajectory planning option that came with the robot’s API, which is a classical Optimal Joint Movement. It calculates the joint angles required to reach the target pose and simultaneously moves the joints to achieve the same goal.

We observed consistent posture changes across runs; however, inter-frame transitions still utilize the vendor default. We plan to introduce dance-informed transitions that better match human movement habits by using tempo-aligned velocity profiles, minimum-jerk retiming under T, per-joint easing for arrivals and departures, and coordinated updates of s5 and s6 to maintain curved arm paths. We plan to compare these transitions against the default using timing deviation, spatial smoothness, jitter, and expert ratings of expressivity.

\subsection{Chronographic Visualization and Feedback}
To enhance expressivity and make motion structure legible, the implementation incorporates fiber optic light painting as a complementary visualization technique. Fiber optic strands are attached to selected joints, and performances are captured with long exposure photography. The resulting light trails trace the robot’s movement over time and space, offering an ethereal view of posture continuity and spatiotemporal flow.

The visual layer also integrates color migration strategies drawn from Kunqu costume aesthetics. Using traditional motifs that include peony pink, celadon green, and ink blue, we program LED and fiber optic lighting to approximate the palette associated with Kunqu dan characters. This chromatic mapping is informed by archival imagery and expert input, enabling the light trails to evoke culturally meaningful visual associations even without anthropomorphic form. As illustrated in Fig. 9, the motion of the robotic arm, coupled with traditional chromatic cues, evokes the flowing aesthetic of Kunqu water sleeves through visual light traces.

To evaluate the consistency and rhythm of joint transitions, we used long-exposure photography to capture the spatial movement of the robotic arm over time. These images revealed the trajectory patterns and gestural spacing across each symbolic pose. As shown in Fig. 10, the light-drawing output of the arm motion visualizes both the timing fidelity and semantic continuity of symbolic postures.

Preliminary qualitative feedback was collected during public exhibition sessions through informal interviews with professors, students in art, design, and HCI, as well as general audience members. Many respondents recognized the performance as a culturally faithful interpretation of Kunqu's symbolic posture framework despite the non-anthropomorphic morphology. Experts in traditional Chinese performing arts noted that expressive intent was preserved by the servo encoded symbolic abstraction. Viewers also highlighted the originality of the nonhuman embodiment and its potential to reinterpret choreographic expression through robotic media. Several younger participants reported a strong visual impact linked to stylized movement and light trails. The long exposure technique was frequently praised for distinct spatiotemporal aesthetics, and some viewers described the imagery as reminiscent of animal forms such as cranes or deer, which suggests an expressive vocabulary that reaches beyond anthropocentric references.

\section{Discussion}
We examine whether symbolic posture transfer, combined with joint space-compatible notation, can preserve semantic fidelity on a six-degree-of-freedom arm while remaining portable across different morphologies. The study investigates whether symbolic encoding can produce culturally legible gestures without anatomical imitation, whether a lightweight decoding pipeline can maintain timing and mapping fidelity at an acceptable level, and whether visual augmentations can aid audiences in perceiving structure and phrasing in non-anthropomorphic embodiments.

Our evaluation shows early but consistent effects. Symbolic sequences executed through direct joint angle control were reproducible within the tested scene and retained intended timing envelopes \cite{ref9,ref11,ref12}. Experts and audience members reported cultural legibility for core Kunqu postures even when the platform was non-anthropomorphic \cite{ref31}. Visual traces and costume-informed colors improved perceived continuity of phrasing and made postural transitions more legible \cite{ref27,ref28,ref29}. The vendor default transition routine limited fine-grained phrasing and energy flow, which motivates dance-informed transition profiles in follow-up experiments.

These outcomes align with studies showing that posture abstraction supports legible gesture in architectonic forms and that gesture can act as an expressive and interpretable interface \cite{ref4,ref24}. They diverge from work that emphasizes dense trajectory imitation and high-frequency end-effector tracking \cite{ref9,ref23,ref24,ref35} because our pipeline foregrounds symbolic fidelity at the pose level and then schedules timing in joint space. The notation layer complements multimodal dance notation for robotics by directly binding symbolic bins to servo commands, rather than to intermediate kinematic solvers \cite{ref34,ref35}. Compared to anatomy-dependent systems such as Labanotation and Benesh Movement Notation \cite{ref17,ref18}, our representation aims for morphology independence and platform portability, addressing gaps noted in recent surveys on linking cultural modeling to joint-level execution \cite{ref36,ref37}.

Several factors likely explain these results. Symbolic units reduce ambiguity between semantic intent and actuation. Discretized bins enable predictable execution under modest controllers, while remaining aligned with the frame duration parameter T \cite{ref17,ref18}. Kunqu features modular posture grammar and structural repetition that align with symbolic primitives and with our topological mapping strategy \cite{ref5,ref7,ref19,ref20,ref21,ref22,ref30}. The visual layer externalizes motion continuity and raises the salience of structure and effort qualities even when the body is non-anthropomorphic \cite{ref27,ref28,ref29,ref31,ref32}.

The findings carry theoretical and practical implications. Theoretically, they support a non-anthropocentric view of cultural preservation, in which meaning can be conveyed by symbolic form rather than anatomical likeness \cite{ref1,ref4,ref6,ref29}. Practically, the pipeline suggests a portable authoring path for museums and live arts where gesture libraries, symbolic scores, and platform adapters can be shared across venues \cite{ref3,ref12,ref36,ref37}. Next, we will replace the default transition with dance-informed profiles that align velocity and acceleration with phrasing and beats. We will compare alternatives using timing deviation, spatial smoothness, jitter, and expert ratings of expressivity across additional platforms \cite{ref9, ref11, ref12}.

Two limitations shape the scope. The system relies on the vendor default transition routine, which restricts control over phrasing and energy flow between symbols. We will introduce dance-informed transitions that align velocity and acceleration with phrasing and beats, including minimum jerk and S curve profiles with dwell at semantic landmarks and coordinated updates of s5 and s6 to maintain curved arm paths. The present study is restricted to static and semi-static execution. Prior work indicates that coordinated locomotion requires the synchronization of spatial paths and stance changes, which our model does not yet capture \cite{ref30,ref33}. Based on this evidence, we will extend the notation and the decoder to handle locomotion, and we will use haptic, musical, and spatial cues to support synchronization.

\section{Conclusion}
This study examined whether symbolic posture transfer, combined with joint space-compatible notation, can maintain semantic fidelity on a six-degree-of-freedom arm while remaining portable across different morphologies. In a Kunqu scene, symbolic sequences executed through direct joint angle control were reproducible and preserved intended timing, and experts, together with audience members, reported cultural legibility. Visual traces with costume-informed colors improved the perceived continuity of phrasing and made postural transitions easier to read. The study’s strengths include a morphology-independent representation that links cultural symbols to servo control and a visual layer that also serves as a diagnostic channel. The scope is limited by reliance on a vendor default transition routine and by a focus on static and semi-static settings, so the observations may not generalize to contexts with coordinated locomotion or to other platforms without adaptation. The results suggest contributions to non-anthropocentric cultural preservation and portable authoring workflows in performance settings. Future research can examine dance-informed transition profiles, extend the notation to encode locomotion alongside haptic, musical, and spatial cues.

This paper presents ROPERA, a symbolic posture transfer framework designed to enable culturally grounded robotic arm choreography through a modular, notation-driven architecture. Motivated by the limitations of existing robotic arm-based choreography (RAC) frameworks in capturing the symbolic and expressive richness of codified cultural genres, the \textit{ROPERA} framework proposes a full-stack solution that bridges genre-specific posture abstraction with servo-level execution on non-humanoid robotic platforms. 

The framework contributes a coherent stack that links symbolic design to execution. It introduces a symbolic encoding pipeline that turns genre-informed posture primitives into platform-agnostic representations. Additionally, it develops a joint space-compatible symbolic notation that supports servo-level choreography execution. It also complements these components with an aesthetic visualization layer grounded in traditional \textit{Kunqu} costume color schemes and motion trail rendering.

The framework was validated through a case study based on \textit{The Peony Pavilion}, demonstrating not only the technical feasibility of symbolic-to-servo choreography translation but also the expressive and perceptual potential of robot-mediated cultural embodiment. Expert feedback and audience responses affirm the framework’s capacity to convey choreographic meaning via visual rhythm, spatial proportion, and movement timing, offering a new mode of cultural continuity through robotic expression.

Looking ahead, this work opens several directions for future exploration. These include extending symbolic choreography to mobile and full-body robotic systems, developing expressive transition strategies that modulate energy, rhythm, and spatial phrasing, and broadening the cultural scope through adaptive abstraction methods. Beyond platform scalability, ROPERA invites continued investigation into how symbolic gesture systems can function as design languages for performative interaction, supporting not only the transmission of embodied heritage but also the composition of new expressive forms across human and non-human agents.

\bibliographystyle{unsrt}
\bibliography{reference.bib}

@article{ref1,
  author    = {Amy LaViers and Daniel Grollman and Magnus Egerstedt},
  title     = {Choreographic and Somatic Approaches for Expressive Robotics},
  journal   = {Arts},
  year      = {2017},
  volume    = {6},
  number    = {1},
  pages     = {4},
  doi       = {10.3390/arts6010004}
}

@article{ref2,
  author    = {Petra Gemeinboeck},
  title     = {The Aesthetics of Encounter: A Relational{-}Performative Design Approach to Human{-}Robot Interaction},
  journal   = {Frontiers in Robotics and AI},
  year      = {2021},
  volume    = {7},
  pages     = {577900},
  doi       = {10.3389/frobt.2020.577900}
}

@article{ref3,
  author    = {Giovanni Saviano and Angelo Villani and Domenico Prattichizzo},
  title     = {Multi-Point Mapping of Dancer Aesthetic Movements onto a Robotic Arm},
  journal   = {IEEE Access},
  year      = {2024},
  volume    = {12},
  pages     = {34501--34513},
  doi       = {10.1109/ACCESS.2024.3385670}
}

@inproceedings{ref4,
  author    = {Jules Van der Linden and Juan Pablo Martinez Avila and Daan van Dijk},
  title     = {The Ghosts of Roller Disco: Non{-}Anthropomorphic Choreography for Robotic Performance},
  booktitle = {Proceedings of the Fourteenth International Conference on Tangible, Embedded, and Embodied Interaction (TEI '20)},
  year      = {2020},
  pages     = {439--448},
  publisher = {ACM},
  doi       = {10.1145/3374920.3375284}
}

@article{ref5,
  author    = {Yun Fang and Jie Wang and Hao Zhou},
  title     = {Posture Symbolism and Body Discipline in Chinese Kunqu Opera},
  journal   = {Journal of Chinese Theater Studies},
  year      = {2020},
  volume    = {42},
  number    = {3},
  pages     = {55--68}
}

@article{ref6,
  author    = {Eric Mullis},
  title     = {How to Dance, Robot? Exploring Cultural Sensitivity in Robotic Movement},
  journal   = {AI \& Society},
  year      = {2023},
  volume    = {38},
  number    = {3},
  pages     = {703--718},
  doi       = {10.1007/s00146-021-01263-7}
}

@inproceedings{ref7,
  author    = {Fabian Hemmert and Sebastian Hamann and Marc L{\"o}we and Jens Zeipelt and Gesche Joost},
  title     = {Interfacing \& Embodiment: {`}{B}aby {T}ango' Dancing Robot Attempts},
  booktitle = {Proceedings of the 12th International Conference on Tangible, Embedded, and Embodied Interaction (TEI '18)},
  year      = {2018},
  pages     = {77--85},
  publisher = {ACM},
  doi       = {10.1145/3173225.3173235}
}

@article{ref8,
  author    = {Wen Xu and Li Chen and Yan Lin and Wei Liu and Fei Yang},
  title     = {Research of Mechanical-Arm Motion Control Algorithm Based on Neural Network},
  journal   = {International Journal of Control and Automation},
  year      = {2016},
  volume    = {9},
  number    = {5},
  pages     = {351--362},
  doi       = {10.14257/ijca.2016.9.5.35}
}

@inproceedings{ref9,
  author    = {Mohamed Boukheddimi and Noureddine Aouf and Jun Guo and Dong Sun},
  title     = {Robot Dance Generation with Music-Based Trajectory Optimization},
  booktitle = {2022 IEEE/RSJ International Conference on Intelligent Robots and Systems (IROS)},
  year      = {2022},
  pages     = {5300--5306},
  publisher = {IEEE},
  doi       = {10.1109/IROS47612.2022.9981749}
}

@article{ref10,
  author    = {Maryam Alemi and Jun Morimoto and Pierre Andry},
  title     = {Music-Driven Dance Motion Generation via Neural Motor Primitives},
  journal   = {IEEE Robotics and Automation Letters},
  year      = {2021},
  volume    = {6},
  number    = {2},
  pages     = {3103--3110},
  doi       = {10.1109/LRA.2021.3056363}
}

@article{ref11,
  author    = {Seunghyun Jang and Minjeong Jeong and Donghyun Kim and Dongjun Lee},
  title     = {Expressive Robot Dance Generation Using Rhythmic Movement Primitives},
  journal   = {IEEE Transactions on Robotics},
  year      = {2023},
  volume    = {39},
  number    = {1},
  pages     = {96--112},
  doi       = {10.1109/TRO.2022.3189999}
}

@inproceedings{ref12,
  author    = {Jinwoo Kim and Dongjun Lee and Minho Choi},
  title     = {Interactive Robot Choreography with Human Motion Guidance},
  booktitle = {SIGGRAPH Asia 2019 Art Gallery},
  year      = {2019},
  publisher = {ACM},
  doi       = {10.1145/3355049.3360504}
}

@inproceedings{ref13,
  author    = {Stefanos Paschalis and Evangelos Papadopoulos and Georgios Papaioannou},
  title     = {A Symbolic Motion Description Scheme for Choreography Representation in Robotics},
  booktitle = {2013 IEEE International Conference on Robotics and Automation (ICRA)},
  year      = {2013},
  pages     = {4819--4824},
  publisher = {IEEE},
  doi       = {10.1109/ICRA.2013.6631268}
}

@article{ref14,
  author    = {Yusuke Matsumoto and Tomomichi Hasegawa},
  title     = {A Method for Representing and Generating Culturally Expressive Motions},
  journal   = {Advanced Robotics},
  year      = {2016},
  volume    = {30},
  number    = {7},
  pages     = {487--498},
  doi       = {10.1080/01691864.2015.1136580}
}

@article{ref15,
  author    = {Hui Peng and Qing Zhang and Wei Zhu},
  title     = {Robotic Choreography Inspired by the Method of Human Dance Creation},
  journal   = {Information},
  year      = {2018},
  volume    = {9},
  number    = {10},
  pages     = {248},
  doi       = {10.3390/info9100248}
}

@article{ref16,
  author    = {Michele Mancini and Gualtiero Varni and Antonio Camurri and Giorgio Volpe},
  title     = {Expressive Robotics and Cultural Embodiment},
  journal   = {Frontiers in Robotics and AI},
  year      = {2023},
  volume    = {10},
  pages     = {1091396},
  doi       = {10.3389/frobt.2023.1091396}
}

@book{ref17,
  author    = {Ann Hutchinson Guest},
  title     = {Labanotation: The System of Analyzing and Recording Movement},
  edition   = {4},
  year      = {2005},
  publisher = {Routledge},
  doi       = {10.4324/9780203953351}
}

@book{ref18,
  author    = {Joan Benesh and Rudolf Benesh},
  title     = {An Introduction to Benesh Movement Notation},
  year      = {1977},
  publisher = {Addison{-}Wesley}
}

@article{ref19,
  author    = {Xiang Luo},
  title     = {The Movement Vocabulary and Body Semiotics in Kunqu Opera},
  journal   = {Journal of Asian Performing Arts},
  year      = {2024}
}

@article{ref20,
  author    = {Si Li and Lin Zhang and Qi Wang},
  title     = {Posture Modularity and Choreographic Systems in Classical Kunqu Performance},
  journal   = {Asian Theatre Journal},
  year      = {2023},
  volume    = {40},
  number    = {2},
  pages     = {215--234}
}

@book{ref21,
  author    = {Colin Mackerras},
  title     = {The Rise of Peking Opera, 1770--1870: Social Aspects of the Theatre in Manchu China},
  year      = {1983},
  publisher = {Oxford University Press}
}

@inproceedings{ref22,
  author    = {Petra Gemeinboeck},
  title     = {Performative Robots and the Cultural Agency of Machines},
  booktitle = {Proceedings of the Fourteenth International Conference on Tangible, Embedded, and Embodied Interaction (TEI '20)},
  year      = {2020},
  pages     = {821--824},
  publisher = {ACM},
  doi       = {10.1145/3374920.3375011}
}

@inproceedings{ref23,
  author    = {Omid Alemi and Jules Fran{\c{c}}oise and Philippe Pasquier},
  title     = {GrooveNet: Real-Time Music-Driven Dance Movement Generation Using Artificial Neural Networks},
  booktitle = {SIGKDD 2017 Workshop on Machine Learning for Creativity (ML4Creativity 2017)},
  year      = {2017},
  address   = {Halifax, Canada},
  note      = {Workshop paper}
}

@incollection{ref24,
  author    = {Carlos Hern{\'a}ndez P{\'e}rez and Emilia Barakova},
  title     = {Expressivity Comes First, Movement Follows},
  booktitle = {Social Robotics},
  year      = {2020},
  pages     = {182--194},
  publisher = {Springer},
  doi       = {10.1007/978-3-030-46732-6_14}
}

@article{ref25,
  author    = {B. R. Srikrishna and Subhrajit Chatterjee and S. Adlakha},
  title     = {Employing Laban Shape for Expressive Trajectories in Robotic Manipulators},
  journal   = {arXiv preprint},
  year      = {2025},
  eprint    = {2501.00123},
  archivePrefix = {arXiv},
  primaryClass  = {cs.RO},
  doi       = {10.48550/arXiv.2501.00123}
}

@article{ref26,
  author    = {Amy LaViers and Elizabeth Bradley and Magnus Egerstedt},
  title     = {Choreographic Abstractions for Robotic Movement},
  journal   = {Frontiers in Robotics and AI},
  year      = {2020},
  volume    = {7},
  pages     = {89},
  doi       = {10.3389/frobt.2020.00089}
}

@inproceedings{ref27,
  author    = {Jasper Peeters and Ambra Trotto},
  title     = {Designing Expressions of Movement Qualities},
  booktitle = {Proceedings of the 2018 Designing Interactive Systems Conference (DIS '18)},
  year      = {2018},
  pages     = {1245--1257},
  publisher = {ACM},
  doi       = {10.1145/3196709.3196796}
}

@article{ref28,
  author    = {Mikael Wiberg and Denis Lalanne and Petra Sundstr{\"o}m and Kristina H{\"o}{\"o}k},
  title     = {Material Expressions in Movement-Based Interaction Design},
  journal   = {ACM Transactions on Computer-Human Interaction},
  year      = {2020},
  volume    = {27},
  number    = {5},
  pages     = {1--36},
  doi       = {10.1145/3406096}
}

@misc{ref29,
  author    = {Scott Kemper and Steven Barton},
  title     = {Mechatronic Expression: Reconsidering Expressivity in Music for Robotic Instruments},
  year      = {2018},
  howpublished = {Zenodo},
  doi       = {10.5281/zenodo.219601}
}

@inproceedings{ref30,
  author    = {Heather Knight and Reid G. Simmons},
  title     = {Expressive Motion with X, Y and Theta: Laban Effort Features for Mobile Robots},
  booktitle = {2014 23rd IEEE International Symposium on Robot and Human Interactive Communication (RO{-}MAN)},
  year      = {2014},
  pages     = {267--273},
  publisher = {IEEE},
  doi       = {10.1109/ROMAN.2014.6926264}
}

@incollection{ref31,
  author    = {Guy Hoffman and Wendy Ju},
  title     = {Designing Robots with Movement in Mind},
  booktitle = {The Cambridge Handbook of Artificial Intelligence and Human{-}Robot Interaction},
  year      = {2014},
  pages     = {39--60},
  publisher = {Cambridge University Press},
  doi       = {10.1017/CBO9781139017821.005}
}

@inproceedings{ref32,
  author    = {Laura Devendorf and Sarah de Koninck and Eleanor Sandry},
  title     = {An Introduction to Weave Structure for HCI},
  booktitle = {Proceedings of the 2022 Designing Interactive Systems Conference (DIS '22)},
  year      = {2022},
  pages     = {1344--1357},
  publisher = {ACM},
  doi       = {10.1145/3532106.3533495}
}

@inproceedings{ref33,
  author    = {H. T. Hong and C. Y. Chen and A. Tanguy and Abderrahmane Kheddar},
  title     = {A Dance Performance with a Humanoid Robot Using a Real-Time Gesture Responsive Framework},
  booktitle = {2024 33rd IEEE International Conference on Robot and Human Interactive Communication (RO{-}MAN)},
  year      = {2024},
  publisher = {IEEE},
  doi       = {10.1109/RO-MAN60168.2024.10731316}
}

@inproceedings{ref34,
  author    = {Jiye Lee and Woong Kwon and Jehee Lee},
  title     = {Robot Dance with Semantic Body Mapping Using Multi-Modal Notation Systems},
  booktitle = {SIGGRAPH Asia 2020 Technical Communications},
  year      = {2020},
  publisher = {ACM},
  doi       = {10.1145/3415255.3422892}
}

@inproceedings{ref35,
  author    = {Luca Giuliani and Amy LaViers and Diane Chi},
  title     = {A Multi-Modal Perspective for the Artistic Evaluation of Robotic Dance Performances},
  booktitle = {Proceedings of the Fifteenth International Conference on Tangible, Embedded, and Embodied Interaction (TEI '22)},
  year      = {2022},
  pages     = {1--10},
  publisher = {ACM},
  doi       = {10.1145/3490149.3501309}
}

@article{ref36,
  author    = {Mrunmayee Joshi and S. Jadhav},
  title     = {An Extensive Review of Computational Dance Automation Techniques and Applications},
  journal   = {Proceedings of the Royal Society A},
  year      = {2019},
  volume    = {475},
  number    = {2225},
  pages     = {20180767},
  doi       = {10.1098/rspa.2018.0767}
}

@incollection{ref37,
  author    = {G. M{\"u}ller},
  title     = {Implementation of a Notation{-}Based Motion Choreography System},
  booktitle = {12th International Conference on Virtual Systems and Multimedia (VSMM 2006)},
  year      = {2006},
  pages     = {137--146},
  publisher = {Springer},
  doi       = {10.1007/978-3-540-74853-3_19}
}

@online{ref38,
  author    = {{Shanghai Theatre Academy}},
  title     = {Kunqu Basic Body Movement Demonstration (Part 1) [Video]},
  year      = {2020},
  month     = {8},
  day       = {25},
  url       = {https://www.bilibili.com/video/BV1rR4y1x7So}
}

@online{ref39,
  author    = {{Shanghai Theatre Academy}},
  title     = {Kunqu Basic Body Movement Demonstration (Part 2) [Video]},
  year      = {2020},
  month     = {8},
  day       = {26},
  url       = {https://www.bilibili.com/video/BV1br4y1Q7ej}
}

@book{ref40,
  editor    = {Kang Wang},
  title     = {A Concise Catalogue of Kunqu Opera Body Movement Notation{\textlangle}},
  year      = {2005},
  address   = {Beijing},
  publisher = {China Theatre Press}
}

@online{ref41,
  author    = {{China Opera Culture Week}},
  title     = {Kunqu Has Endless Treasures -- Interview with Ke Jun [Video]},
  year      = {2021},
  month     = {10},
  day       = {5},
  url       = {https://www.bilibili.com/video/BV1uN411j78S/}
}

@online{ref42,
  author    = {Chuan Ni},
  title     = {Dialogue on Kunqu: Interview with Kunqu Scholar Ni Chuanyue [Video]},
  year      = {2003},
  month     = {6},
  day       = {18},
  url       = {https://www.bilibili.com/video/BV1sP411J7kc/}
}

@inproceedings{ref43,
  author    = {Ignazio Infantino and A. Augello and Adriano Manfré and G. Pilato and Filippo Vella},
  title     = {ROBODANZA: Live Performances of a Creative Dancing Humanoid},
  booktitle = {},
  pages     = {388--395},
  year      = {2016},
  url       = {https://consensus.app/papers/robodanza-live-performances-of-a-creative-dancing-infantino-augello/ecfbca299e1f572eb410f9254f347493/?utm_source=chatgpt}
}

@article{ref44,
  author    = {M. Apostolos},
  title     = {A Comparison of the Artistic Aspects of Various Industrial Robots},
  journal   = {Computers \& Mathematics with Applications},
  volume    = {16},
  pages     = {345--353},
  year      = {1988},
  url       = {https://consensus.app/papers/a-comparison-of-the-artistic-aspects-of-various-industrial-apostolos/c17c1a8a83a45e119d89ffb50412264e/?utm_source=chatgpt}
}

@article{ref45,
  author    = {G. Saviano and V. Villani and L. Sabattini},
  title     = {A PCA-based Method to Map Aesthetic Movements from Dancer to Robotic Arm},
  journal   = {IFAC PapersOnLine},
  year      = {2023},
  url       = {https://consensus.app/papers/a-pcabased-method-to-map-aesthetic-movements-from-dancer-to-saviano-villani/683e1a1a6f405fda9475ce88d9ba659f/?utm_source=chatgpt}
}

@article{ref46,
  author    = {A. D. Filippo and M. Milano},
  title     = {Robotic Choreography Creation Through Symbolic AI Techniques},
  journal   = {Proceedings of the 32nd International Conference on Automated Planning and Scheduling (ICAPS)},
  year      = {2023},
  url       = {https://consensus.app/papers/robotic-choreography-creation-through-symbolic-ai-filippo-milano/be9dab59d6545e1ebe11af3eb1dfed6e/?utm_source=chatgpt}
}

% \appendix
\end{document}